\documentclass[11pt,a4paper]{article}

\usepackage[utf8]{inputenc}
\usepackage[T1]{fontenc}
\usepackage{times}
\usepackage{amsmath,amssymb,amsfonts}
\usepackage{graphicx}
\usepackage{booktabs}
\usepackage{hyperref}
\usepackage{xcolor}
\usepackage{algorithm}
\usepackage{algorithmic}
\usepackage[margin=2.5cm]{geometry}
\usepackage{caption}
\usepackage{subcaption}
\usepackage{tikz}
\usepackage{pgfplots}
\pgfplotsset{compat=1.18}
\usetikzlibrary{arrows.meta,positioning,shapes.geometric,fit,backgrounds}

\hypersetup{
    colorlinks=true,
    linkcolor=blue!60!black,
    citecolor=blue!60!black,
    urlcolor=blue!60!black
}

\title{\textbf{MAGNET: Autonomous Expert Model Generation\\via Decentralized Autoresearch and BitNet Training}}

\author{
    Yongwan Kim\textsuperscript{*} \quad Sungchul Park\textsuperscript{*}\\[4pt]
    Holo Studio Co., Ltd.\\
    \texttt{https://holostudio.io}\\[2pt]
    {\small \textsuperscript{*}Equal contribution. Correspondence: Y.~Kim (CTO), S.~Park (CEO)}
}

\date{March 2026}

\begin{document}

\maketitle

\begin{abstract}
We present MAGNET (\textbf{M}odel \textbf{A}utonomously \textbf{G}rowing \textbf{Net}work), a decentralized system for autonomous generation, training, and serving of domain-expert language models across commodity hardware. MAGNET integrates four components: (1)~\emph{autoresearch}, an autonomous ML research pipeline that automates dataset generation, hyperparameter exploration, evaluation, and error-driven iteration; (2)~BitNet b1.58 ternary training, enabling CPU-native inference via bitnet.cpp without GPU hardware; (3)~DiLoCo-based distributed merging for communication-efficient aggregation of domain specialists; and (4)~on-chain contribution tracking on the HOOTi EVM chain. Inspired by Karpathy's iterative research methodology~\cite{karpathy2019, karpathy2026}, MAGNET automates and decentralizes the experiment-failure-hypothesis cycle.

We validate autoresearch through three case studies: video safety classification (Zevor: balanced accuracy $0.9287 \rightarrow 0.9851$, false negatives $9 \rightarrow 0$, across 5{,}000+ configurations), cryptocurrency directional prediction (StockClaw: $41\% \rightarrow 54.9\%$ hit rate across three versions), and BitNet hyperparameter optimization (Genkidama: 10-phase sweep, $-16.7\%$ validation loss). We further demonstrate partial end-to-end validation through Genkidama, a 618M-parameter BitNet b1.58 model whose pretraining is complete (480{,}000 steps, best validation loss 2.3762) with verified GGUF export for bitnet.cpp CPU inference; ORPO conversational alignment is in progress.

\textbf{Validation status.} Autoresearch is empirically validated (three case studies). BitNet training is empirically validated (Genkidama 618M). The on-chain incentive module is implemented and unit-tested (daily emission cap, commit-reveal, differentiated slashing, bootstrap notary consensus) but not yet deployed on a public mainnet. DiLoCo distributed merging is designed but not yet experimentally validated. Our primary contribution is the autoresearch methodology and its empirical validation; the system-level integration of all four components into a decentralized research architecture has not, to our knowledge, been previously described in the literature.
\end{abstract}

\section{Introduction}

The current landscape of large language models is characterized by extreme centralization. Training a frontier model requires thousands of GPUs, proprietary datasets, and engineering teams that only a handful of corporations can assemble. While frameworks like TensorFlow and PyTorch have \emph{democratized model training}, the \emph{research process} itself (data curation, architecture search, hyperparameter optimization, evaluation) remains a manual, expert-driven bottleneck.

Karpathy's ``A Recipe for Training Neural Networks''~\cite{karpathy2019} laid out a principled, sequential methodology for ML practitioners: inspect data carefully, build an evaluation skeleton, overfit a single batch, regularize, tune, and squeeze out remaining gains. The blog emphasized paranoid, incremental validation at each stage and the primacy of data quality; in Karpathy's words, ``the neural net is effectively a compressed/compiled version of your dataset.'' More recently, Karpathy's \texttt{autoresearch} project~\cite{karpathy2026} demonstrated that a compact script ($\sim$630 lines of Python) can autonomously run ML experiments (generating hypotheses, executing training runs, evaluating results, and iterating), embodying the principle that the research loop itself can be automated.

MAGNET's autonomous research pipeline, which we call ``autoresearch''\footnote{We use the term ``autoresearch'' as a descriptive shorthand for autonomous research loops. Karpathy's \texttt{autoresearch} project~\cite{karpathy2026} independently uses the same name; our usage is not a claim of priority.}, extends this concept into a decentralized setting. Rather than a single script on a single machine, MAGNET distributes the autonomous research loop across a network of nodes. Each node independently runs error-driven iterations (data generation, training, evaluation, strategy pivots) while the network aggregates results through DiLoCo merging. The core loop structure is preserved; what changes is the scale and the removal of any centralized coordinator.

We identify four fundamental requirements for truly decentralized AI \emph{research}:

\begin{enumerate}
    \item \textbf{Research autonomy}: The research loop must run autonomously (data generation, architecture search, hyperparameter optimization, evaluation, and error-driven iteration) without human intervention.
    \item \textbf{Hardware accessibility}: Trained models must be \emph{servable} on commodity CPU nodes via ternary quantization; training is distributed across GPU-capable nodes via DiLoCo, but inference participation requires no GPU.
    \item \textbf{Knowledge aggregation}: Independently trained models must be mergeable into collectively stronger models.
    \item \textbf{Incentive integrity}: Contributions must be verifiable and rewards tamper-proof.
\end{enumerate}

These four requirements map directly to MAGNET's pillars. BitNet b1.58 (Section~4) addresses hardware accessibility by enabling CPU-native inference via ternary weights. Autoresearch (Section~5) provides research autonomy through error-driven iteration. DiLoCo (Section~10) enables knowledge aggregation via communication-efficient distributed merging. On-chain incentives (Section~13) address incentive integrity. The following sections describe each pillar, with autoresearch validated through three case studies (Sections~6--8).

Existing systems address important subsets of these requirements (Section~2). MAGNET integrates all four within a single architecture; we present the design, validate three pillars with working implementations (autoresearch, BitNet training, DiLoCo merging), and report initial on-chain deployment on a test chain.

\textbf{Open-source commitment.}\quad MAGNET's core research pipeline (autoresearch, BitNet training, model export, and evaluation) is planned for public open-source release\footnote{Repository forthcoming at \url{https://github.com/holostudio/magnet-core}. The codebase is functional internally; public release is pending documentation and license finalization.}. The architecture follows a core/platform separation: the \emph{core} (research methodology, training code, model serving) will be open for anyone to fork, extend, and deploy, while the \emph{platform} layer (on-chain rewards, node coordination, incentive distribution) operates on the HOOTi network.
\subsection{Contributions}

\begin{itemize}
    \item A formal description of the four-pillar architecture and analysis of component interdependencies (Section~\ref{sec:architecture}).
    \item Integration of BitNet b1.58 ternary training with LLaMA-compatible architecture for bitnet.cpp CPU inference (Section~\ref{sec:bitnet}).
    \item \textbf{Autoresearch}: An autonomous research pipeline validated across three production domains with quantitative case studies (Sections~6--8).
    \item A DiLoCo-based protocol for merging heterogeneous domain specialists (Section~\ref{sec:diloco}).
    \item An on-chain incentive mechanism with commit-reveal contribution proofs, implemented and unit-tested on the HOOTi EVM chain (Section~\ref{sec:onchain}).
    \item Genkidama: a 618M-parameter BitNet b1.58 model demonstrating partial end-to-end validation (autoresearch configuration $\to$ pretraining $\to$ HuggingFace export $\to$ bitnet.cpp CPU inference; ORPO alignment in progress).
\end{itemize}

\section{Related Work}

\textbf{Decentralized ML.}\quad Federated Learning~\cite{mcmahan2017} distributes training across decentralized clients, including mobile and edge devices, by exchanging model updates rather than raw data. Unlike MAGNET, it does not target autonomous research loops, ternary-weight deployment, or on-chain incentive design. Bittensor~\cite{bittensor2021} is a decentralized incentive network whose ecosystem spans inference, training, data curation, and RLHF across multiple subnets; recent work~\cite{gauntlet2025} explores permissionless distributed LLM training with token-based rewards on the Bittensor chain. MAGNET differs in centering autonomous research iteration and BitNet-based specialist-to-generalist merging. Gensyn provides a protocol for executing, verifying, and coordinating ML tasks on arbitrary devices, with the NoLoCo protocol~\cite{noloco2025} for communication-efficient training. Prime Intellect~\cite{primeintellect2024} trained INTELLECT-1 (10B parameters) across five countries using OpenDiLoCo, demonstrating large-scale decentralized pretraining. LLM-Net~\cite{llmnet2025} proposes blockchain-based expert networks with performance verification and accountability.

\textbf{Efficient architectures.}\quad BitNet~\cite{wang2023bitnet} introduced 1-bit training; BitNet b1.58~\cite{ma2024} extended this to ternary weights $\{-1, 0, +1\}$ matching full-precision quality. bitnet.cpp~\cite{bitnetcpp2024} provides optimized CPU kernels for ternary inference. Spectra~\cite{kaushal2024} demonstrated that ternary models exhibit superior per-bit scaling efficiency, remaining effective at high token-to-parameter ratios, suggesting that smaller ternary models with sufficient data can be surprisingly competitive.

\textbf{Distributed training.}\quad DiLoCo~\cite{douillard2024} demonstrates that distributed training with infrequent gradient synchronization (inner steps $H{=}500$, outer optimizer: SGD with Nesterov momentum) matches centralized quality while communicating 500$\times$ less. However, Acker et al.~\cite{diloco_drift2025} found that DiLoCo-pretrained weights may suffer from irreversible representation drift that impairs downstream alignment, a limitation we acknowledge for MAGNET's instruction-tuning stage.

\textbf{Automated research.}\quad Karpathy~\cite{karpathy2019} articulated a principled sequential methodology for neural network training, emphasizing incremental validation and data primacy. His later \texttt{autoresearch} project~\cite{karpathy2026} demonstrated that a single script can autonomously run ML experiments end-to-end. Concurrently, Liu et al.~\cite{autoresearch_vision2025} proposed a vision for fully automated research using LLM agents across the entire scientific pipeline. AutoML~\cite{hutter2019} automates hyperparameter optimization and architecture search but operates within fixed search spaces in centralized settings. Self-Instruct~\cite{wang2023selfinst} and Alpaca~\cite{taori2023} use LLMs for one-shot data generation but do not form continuous autonomous loops. MAGNET's autoresearch pipeline extends these ideas into a decentralized, multi-version pipeline with error-driven strategy pivots and convergence detection.

\section{System Architecture}
\label{sec:architecture}

MAGNET consists of four subsystems in cyclic dependency:

\begin{figure}[h]
\centering
\begin{tikzpicture}[
    box/.style={draw, rounded corners=4pt, minimum width=3.2cm, minimum height=1.4cm, align=center, font=\small\bfseries, line width=0.8pt},
    arr/.style={-{Stealth[length=6pt]}, line width=1pt, color=black!70},
    node distance=2.8cm
]
    \node[box, fill=red!10] (ar) {Autoresearch\\{\footnotesize Data + Config}};
    \node[box, fill=blue!10, right=of ar] (bn) {BitNet b1.58\\{\footnotesize Ternary Training}};
    \node[box, fill=green!10, below=of bn] (dl) {DiLoCo\\{\footnotesize Distributed Merge}};
    \node[box, fill=orange!10, below=of ar] (oc) {On-Chain\\{\footnotesize Incentive Loop}};

    \draw[arr] (ar) -- node[above, font=\scriptsize] {training data} (bn);
    \draw[arr] (bn) -- node[right, font=\scriptsize] {checkpoints} (dl);
    \draw[arr] (dl) -- node[above, font=\scriptsize] {contributions} (oc);
    \draw[arr] (oc) -- node[left, font=\scriptsize] {rewards $\to$ nodes} (ar);

    \draw[arr, dashed, color=red!50!black] (dl.south west) .. controls +(-1.2,-0.5) and +(1.2,-0.5) .. node[below, font=\scriptsize, color=red!50!black] {merged model} (oc.south east);
\end{tikzpicture}
\caption{MAGNET four-pillar architecture. Arrows indicate data flow between subsystems.}
\label{fig:architecture}
\end{figure}
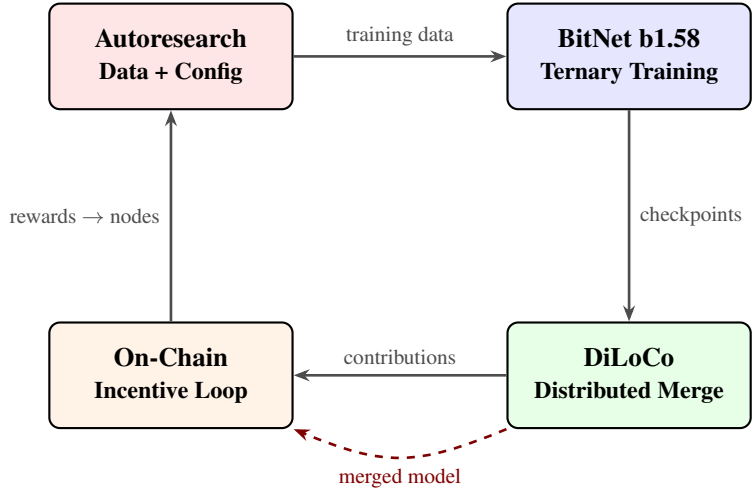

\begin{table}[h]
\centering
\caption{Component interdependencies: degradation when each subsystem is removed. Alternatives exist for each (4-bit quantization for BitNet, FedAvg for DiLoCo, reputation systems for on-chain), but each alternative weakens a specific property that MAGNET optimizes for.}
\label{tab:completeness}
\begin{tabular}{@{}lp{6cm}p{3cm}@{}}
\toprule
\textbf{Without\ldots} & \textbf{Degradation} & \textbf{Alternative exists?} \\
\midrule
Autoresearch & Human bottleneck limits domains and update frequency. & Manual research \\
BitNet & GPU cost excludes most nodes. 4-bit quant is an option but less efficient on CPU. & GPTQ, AWQ \\
DiLoCo & Models remain isolated. FedAvg possible but less communication-efficient. & FedAvg, DeMo \\
On-Chain & Contributions hard to verify at scale. Reputation systems partially substitute. & Reputation, payments \\
\bottomrule
\end{tabular}
\end{table}

\section{BitNet b1.58: Ternary Training for Commodity Hardware}
\label{sec:bitnet}

Having described the four-pillar architecture, we now detail the technical foundation that makes MAGNET accessible to commodity hardware.

Decentralization benefits from broad hardware accessibility. BitNet b1.58 ternary weights $\{-1, 0, +1\}$ enable \textbf{CPU-native inference} via bitnet.cpp, allowing any node to \emph{serve} models without GPU. Note that \emph{training} still benefits from GPU acceleration; MAGNET's hardware accessibility claim applies primarily to the inference/serving layer, while training is distributed across GPU-capable nodes via DiLoCo.

\subsection{Architecture}

We adopt LLaMA-compatible architecture to ensure bitnet.cpp compatibility:

\begin{itemize}
    \item SwiGLU MLP (gate\_proj, up\_proj, down\_proj)
    \item RMSNorm (pre-norm)
    \item Grouped Query Attention with RoPE
    \item BitLinear layers with Straight-Through Estimator (STE)
\end{itemize}

Weight quantization: $\tilde{W} = \text{RoundClamp}(W / \alpha, -1, +1)$, where $\alpha = \text{mean}(|W|)$.

Activation quantization: $\tilde{X} = \text{Quant}(X, 127 / \max(|X|))$.

\subsection{The Python Inference Bottleneck}

A key motivation for BitNet + bitnet.cpp is eliminating Python's FFI overhead in inference loops. When inference is iterated (agent loops, world models, tree search), Python's per-call overhead compounds:

\begin{table}[h]
\centering
\caption{Python FFI overhead as a fraction of total inference time per call. For small models, overhead becomes non-negligible; over $N$ iterations total wasted time is $N \times t_{\text{python}}$.}
\label{tab:python-overhead}
\begin{tabular}{@{}lccc@{}}
\toprule
\textbf{Model} & $t_{\text{infer}}$ & $t_{\text{python}}$ & \textbf{Overhead ratio} $\frac{t_{\text{python}}}{t_{\text{infer}} + t_{\text{python}}}$ \\
\midrule
175B (GPT-3) & $\sim$10\,s & $\sim$5\,$\mu$s & $\sim$0.00005\% \\
7B (Llama 2) & $\sim$500\,ms & $\sim$5\,$\mu$s & $\sim$0.001\% \\
618M (BitNet) & $\sim$10\,ms & $\sim$5\,$\mu$s & $\sim$0.05\% \\
100M (Edge) & $\sim$1\,ms & $\sim$5\,$\mu$s & $\sim$0.5\% \\
\bottomrule
\end{tabular}
\end{table}

The overhead ratio is $t_{\text{python}} / (t_{\text{infer}} + t_{\text{python}})$: the fraction of each call spent in Python FFI rather than inference. While individually small for large models, at the 100M--618M scale this overhead compounds across agent loops (e.g., 50 iterations $\times$ 5\,$\mu$s = 250\,$\mu$s wasted per loop), motivating a single-process C++ design.

bitnet.cpp serves inference and orchestration in the same C++ process with zero FFI overhead, critical for decentralized nodes running agent reasoning loops on commodity CPUs.

\subsection{GGUF Export Pipeline}

\vspace{-0.5em}
\begin{center}
PyTorch (float32 latent) $\xrightarrow{\text{export\_hf()}}$ HuggingFace format $\xrightarrow{\text{converter}}$ GGUF $\xrightarrow{\text{bitnet.cpp}}$ CPU inference
\end{center}
\vspace{-0.5em}

All tensor names match bitnet.cpp's expected schema, enabling direct conversion.

\section{Autoresearch: Decentralized Autonomous ML Research}
\label{sec:autoresearch}

With BitNet enabling any CPU node to serve models, the remaining bottleneck is the research process itself. Autoresearch addresses this by extending the concept demonstrated by Karpathy's \texttt{autoresearch} project~\cite{karpathy2026} (a compact script that autonomously runs ML experiments) into a decentralized, multi-version pipeline. Where Karpathy's implementation runs on a single machine with a fixed experimental scope, MAGNET's autoresearch operates across distributed nodes, adaptively expands its search space based on error analysis, and integrates convergence detection to avoid wasted compute.

Unlike AutoML~\cite{hutter2019}, which optimizes within a fixed search space (e.g., hyperparameter ranges), autoresearch adaptively expands the search space based on error analysis. When a version hits a ceiling, autoresearch generates fundamentally new strategies (architectural pivots, feature engineering changes, or ensemble methods), mirroring how a human researcher would respond to a plateau. This error-driven strategy generation is what distinguishes autoresearch from grid search or Bayesian optimization.

\subsection{Core Loop}

\begin{algorithm}[h]
\caption{Autoresearch Core Loop}
\label{alg:autoresearch}
\begin{algorithmic}[1]
\STATE \textbf{Input}: Domain $D$, evaluation set $\mathcal{E}$, teacher model $T$
\STATE $\mathcal{K} \leftarrow \emptyset$ \COMMENT{Knowledge base}
\STATE $M_0 \leftarrow$ baseline model or pipeline
\FOR{version $v = 1, 2, \ldots$}
    \STATE $\mathcal{F}_v \leftarrow \text{ErrorAnalysis}(M_{v-1}, \mathcal{E})$ \COMMENT{Identify failure modes}
    \STATE $\mathcal{S}_v \leftarrow \text{DesignStrategies}(\mathcal{F}_v, \mathcal{K})$ \COMMENT{Generate fix strategies}
    \FOR{each strategy $s \in \mathcal{S}_v$}
        \STATE $\mathcal{C}_s \leftarrow \text{ConfigSweep}(s)$ \COMMENT{Generate configurations}
        \FOR{each config $c \in \mathcal{C}_s$}
            \STATE Train model $m_c$ with config $c$
            \STATE Evaluate $m_c$ on $\mathcal{E}$
        \ENDFOR
    \ENDFOR
    \STATE $M_v \leftarrow \arg\max_{m_c} \text{score}(m_c, \mathcal{E})$
    \STATE $\mathcal{K} \leftarrow \mathcal{K} \cup \text{Insights}(v)$ \COMMENT{Update knowledge base}
    \IF{$\text{score}(M_v) - \text{score}(M_{v-1}) < \varepsilon$ for $p$ consecutive versions}
        \STATE \textbf{stop} \COMMENT{Convergence detected}
    \ENDIF
\ENDFOR
\RETURN $M_v$, $\mathcal{K}$
\end{algorithmic}
\end{algorithm}

\noindent\textbf{Note on evaluation set reuse.}\quad Algorithm~\ref{alg:autoresearch} uses the same evaluation set $\mathcal{E}$ for error analysis, model selection, and stopping. Reusing $\mathcal{E}$ across many versions risks adaptive overfitting, where the selected model implicitly specializes to $\mathcal{E}$. In our case studies, we mitigate this via leave-one-context-out (LOCO) validation and bootstrap confidence intervals (Section~\ref{sec:case-studies}). A held-out test set, separate from $\mathcal{E}$, should be reserved for final reporting; when this is infeasible (e.g., small proprietary datasets), the adaptive overfitting risk should be acknowledged as a limitation.

\subsection{Key Design Principles}

\textbf{Error-driven iteration.}\quad Each version begins with error analysis of the previous best model, identifying systematic failure patterns. Strategies are designed specifically to address these patterns, not random search.

\textbf{Strategy diversity.}\quad Each version explores 3--8 fundamentally different approaches (e.g., ensemble methods, feature engineering, model architecture changes, threshold optimization), ensuring the search escapes local optima.

\textbf{Convergence detection.}\quad When $p$ consecutive versions each improve by less than $\varepsilon$, autoresearch declares a hard ceiling and terminates, preventing wasted compute.

\textbf{Knowledge accumulation.}\quad Insights from each version (e.g., ``feature noise above 60 features hurts,'' ``per-context thresholds overfit on small subgroups'') are accumulated and inform future versions. This makes practitioner intuition explicit and transferable across nodes.

\textbf{Decentralizability.}\quad Each step of the autoresearch loop is self-contained: a node needs only its local data, the current best model, and the accumulated knowledge base. No centralized coordinator is required during the loop execution. Nodes submit their final results (trained model + knowledge base updates) to the network for DiLoCo aggregation.

\section{Case Study 1: Zevor, Video Safety Classification}
\label{sec:case-studies}

\subsection{Problem}

Zevor (formerly Holo AiD Guardian) is a video safety analysis platform developed internally at Holo Studio, using a multi-signal pipeline: CLIP Safety Head (98.65\% frame accuracy), UCF Violence detector, and Context Classifier. The final SAFE/UNSAFE decision is made by 350 lines of hand-tuned rules achieving $\text{bal\_acc} = 0.9499$ on a 592-video evaluation set (FP=17, FN=11).

\subsection{Failed Approach: End-to-End Transformer Classifier}

The natural first instinct was to replace the hand-tuned rules entirely with a learned model. We trained an end-to-end transformer classifier that consumed the same multi-signal inputs (CLIP embeddings, violence scores, context features) and directly predicted SAFE/UNSAFE. The motivation was clear: a learned model could automatically capture complex decision boundaries that 350 lines of if/else logic might miss, and would be maintainable as the pipeline evolved.

After 690+ experiments with various architectures, training configurations, and regularization strategies, the best transformer achieved only $\text{bal\_acc} = 0.9287$, which is \emph{below} the rule-based pipeline it was intended to replace. The root cause was data scarcity: 592 video-level samples are far too few for a transformer to learn the nuanced multi-signal decision surface that human experts had painstakingly encoded in rules over months.

This failure motivated the key insight behind autoresearch's approach: the problem was not the data size, but the \emph{scope} of what the model was asked to learn.

\subsection{Autoresearch Solution: Decision Head}

The transformer failed because it attempted to learn the \emph{entire} decision surface from raw multi-signal inputs, effectively re-deriving from 592 samples the domain knowledge that experts had encoded over months. Autoresearch inverted the approach: keep the pipeline's signal extraction (CLIP, violence detection, context classification) intact, and replace \emph{only} the 350-line rule-based decision logic with a learned classifier. This dramatically reduced the learning target: the model no longer needed to learn \emph{what} each signal means, only \emph{how to combine} pre-computed signals that already encode rich domain knowledge.

The key breakthrough was \textbf{meta-learning}: including the pipeline's own prediction as a feature transformed the problem from ``classify this video from raw signals'' to ``when does the pipeline get it wrong?'' This is a far simpler function to approximate, and 592 samples proved more than sufficient. In practice, tree-based models (XGBoost + ExtraTrees) achieved 0.9851 balanced accuracy where the transformer had plateaued at 0.9287.

Crucially, this pivot was not designed upfront but \emph{discovered} through the autoresearch loop itself, embodying the iterative research cycle advocated by Karpathy~\cite{karpathy2026}: run an experiment, analyze failures, form a hypothesis, and iterate. The 690 transformer experiments were not wasted; their failure pattern (high variance on small data, inability to surpass a rule-based system) directly informed the hypothesis that the learning scope, not the data budget, was the bottleneck. Autoresearch automated exactly this cycle, enabling the system to identify the practical trade-off (preserve domain-encoded signals, learn only the decision boundary) that a human researcher might have taken weeks to reach.

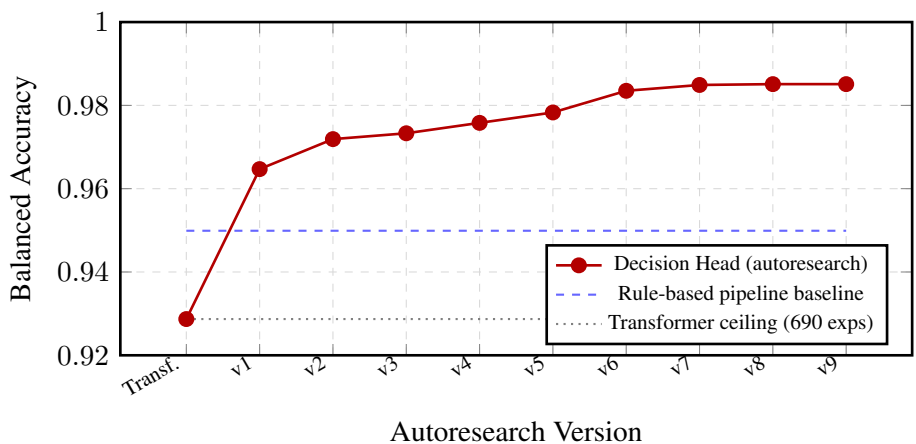
\begin{figure}[h]
\centering
\begin{tikzpicture}
\begin{axis}[
    width=12cm, height=6cm,
    xlabel={Autoresearch Version},
    ylabel={Balanced Accuracy},
    xtick={0,1,2,3,4,5,6,7,8,9},
    xticklabels={Transf., v1, v2, v3, v4, v5, v6, v7, v8, v9},
    xticklabel style={rotate=30, anchor=east, font=\scriptsize},
    ymin=0.92, ymax=1.0,
    grid=major,
    grid style={dashed, gray!30},
    legend pos=south east,
    legend style={font=\scriptsize},
    mark size=2.5pt,
    line width=1pt,
]
\addplot[color=red!70!black, mark=*] coordinates {
    (0, 0.9287) (1, 0.9647) (2, 0.9719) (3, 0.9733)
    (4, 0.9758) (5, 0.9783) (6, 0.9835) (7, 0.9849)
    (8, 0.9851) (9, 0.9851)
};
\addlegendentry{Decision Head (autoresearch)}

\addplot[color=blue!60, dashed, line width=0.8pt] coordinates {
    (0, 0.9499) (9, 0.9499)
};
\addlegendentry{Rule-based pipeline baseline}

\addplot[color=gray, dotted, line width=0.8pt] coordinates {
    (0, 0.9287) (9, 0.9287)
};
\addlegendentry{Transformer ceiling (690 exps)}

\end{axis}
\end{tikzpicture}
\caption{Zevor autoresearch progression. 9 versions, $\sim$5{,}000 configurations. Each version addresses specific failure modes identified in the previous version's error analysis.}
\label{fig:zevor}
\end{figure}

\begin{table}[h]
\centering
\caption{Zevor autoresearch progression: key innovations per version.}
\label{tab:zevor-versions}
\begin{tabular}{@{}clcccp{4.5cm}@{}}
\toprule
\textbf{Ver.} & \textbf{bal\_acc} & \textbf{FP} & \textbf{FN} & \textbf{Configs} & \textbf{Key Innovation} \\
\midrule
Transf. & 0.9287 & 38 & 9 & 690 & End-to-end transformer (failed) \\
v1 & 0.9647 & 24 & 1 & $\sim$100 & Tree models + pipeline\_pred as feature \\
v2 & 0.9719 & 12 & 5 & $\sim$200 & ExtraTrees + class weights \\
v3 & 0.9733 & 13 & 4 & $\sim$150 & Per-context LE specialist \\
v4 & 0.9758 & 11 & 4 & $\sim$200 & Error analysis + model combination \\
v5 & 0.9783 & 9 & 4 & 298 & Context-conditioned thresholds \\
v6 & 0.9835 & 9 & 2 & 1{,}298 & XGBoost + ensemble synergy \\
v7 & 0.9849 & 10 & 1 & 1{,}615 & Fine-tuning + LOCO validation \\
v8 & \textbf{0.9851} & 12 & \textbf{0} & 222 & Per-context probability offsets \\
v9 & 0.9851 & 12 & 0 & 197 & Hard ceiling confirmed \\
\bottomrule
\end{tabular}
\end{table}

\subsection{Results}

The final model (XGBoost 30\% + ExtraTrees 70\% ensemble with per-context probability offsets, top-40 features) achieves:

\begin{itemize}
    \item $\text{bal\_acc} = 0.9851$ ($+3.5\%$ over pipeline, $+5.6\%$ over transformer baseline)
    \item \textbf{Zero false negatives} (11 $\rightarrow$ 0, 100\% elimination)
    \item LOCO validation: 0.9851 (zero overfitting gap)
    \item Bootstrap 95\% CI: [0.9774, 0.9944]
    \item Total errors: 28 $\rightarrow$ 12 (57\% reduction)
\end{itemize}

\textbf{Critical insight}: Per-context \emph{thresholds} overfit on small subgroups (LOCO gap = 0.0204). Per-context \emph{probability offsets} through a single global threshold provide natural regularization (LOCO gap = 0.0000).

\textbf{Caveats.}\quad Zevor results are from an internal 592-video evaluation set with internal labeling. While LOCO and bootstrap validation provide robustness checks, independent reproduction would require access to the proprietary video dataset and pipeline. We present this as an internal case study demonstrating autoresearch's iterative capabilities.

\section{Case Study 2: StockClaw, Cryptocurrency Directional Prediction}

\subsection{Problem}

StockClaw is a cryptocurrency directional prediction system developed internally at Holo Studio, predicting 4-hour price direction (BUY/SELL/HOLD) for 6 major tokens (BTC, ETH, SOL, AVAX, DOGE, XRP). The system initially used ORPO-trained LLM (Qwen3-8B + LoRA) for both report generation and directional judgment.

\subsection{LLM Judgment Failure}

While the ORPO-trained LLM produced high-quality analytical reports, its directional \emph{judgment} accuracy was poor:

\begin{table}[h]
\centering
\caption{StockClaw: LLM-only vs ML judgment comparison.}
\label{tab:stockclaw-llm-ml}
\begin{tabular}{@{}lccc@{}}
\toprule
\textbf{System} & \textbf{F1} & \textbf{Bal. Accuracy} & \textbf{Accuracy} \\
\midrule
ORPO LLM only & 0.173 & 0.269 & 0.232 \\
LGBM + Base LLM & 0.371 & 0.371 & 0.374 \\
LGBM + ORPO LLM & 0.372 & 0.371 & 0.374 \\
\textbf{LGBM alone (no LLM)} & \textbf{0.380} & \textbf{0.379} & \textbf{0.384} \\
\bottomrule
\end{tabular}
\end{table}

Key finding: \textbf{Adding the LLM to ML judgment provides zero benefit}; the ML model alone outperforms all LLM-augmented variants. The LLM's value is confined to report quality, not numerical prediction.

\subsection{Autoresearch Solution: From-Scratch Judgment ML}

Autoresearch built a pure ML judgment system through three iterative versions:

\begin{figure}[h]
\centering
\begin{tikzpicture}
\begin{axis}[
    width=12cm, height=5.5cm,
    xlabel={Autoresearch Version},
    ylabel={Hit Rate (\%)},
    xtick={4,5,6},
    xticklabels={v4 (14 days), v5 (1 year), v6 (6.5 years)},
    ymin=35, ymax=60,
    grid=major,
    grid style={dashed, gray!30},
    nodes near coords,
    nodes near coords style={font=\scriptsize, above},
    every node near coord/.append style={yshift=3pt},
    bar width=1.2cm,
    ybar,
]
\addplot[fill=red!60!black, draw=red!80!black] coordinates {
    (4, 41.0) (5, 51.79) (6, 54.90)
};
\end{axis}
\end{tikzpicture}
\caption{StockClaw hit rate progression across autoresearch versions. Data scale: 504 $\rightarrow$ 11{,}904 $\rightarrow$ 77{,}771 samples.}
\label{fig:stockclaw}
\end{figure}
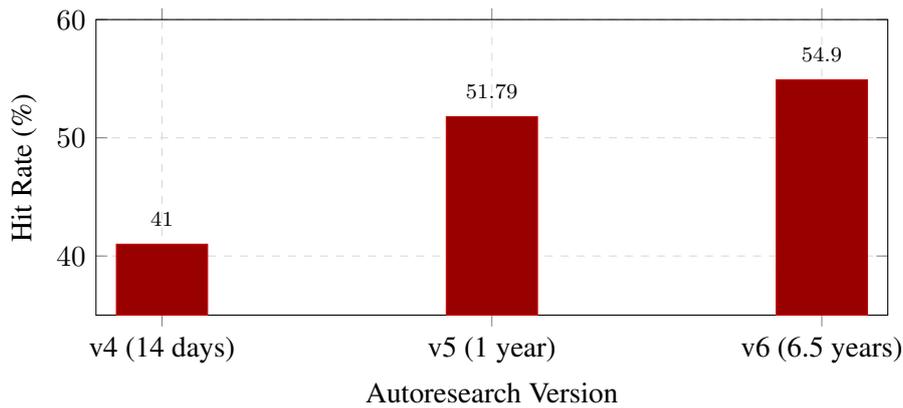

\begin{table}[h]
\centering
\caption{StockClaw autoresearch progression.}
\label{tab:stockclaw-versions}
\begin{tabular}{@{}lccccl@{}}
\toprule
\textbf{Ver.} & \textbf{Data Range} & \textbf{Samples} & \textbf{Features} & \textbf{Hit Rate} & \textbf{Key Innovation} \\
\midrule
v4 & 14 days & 504 & 82+6 & 41.0\% & Futures data integration (+8.3\%) \\
v5 & 1 year & 11{,}904 & 82+7 & 51.8\% & Extended history, funding rate accum. \\
v6 & 6.5 years & 77{,}771 & 82+10 & \textbf{54.9\%} & Full market cycle coverage \\
\bottomrule
\end{tabular}
\end{table}

\subsection{Results}

\begin{itemize}
    \item Hit rate: $41.0\% \rightarrow 54.9\%$ ($+13.9$ pp across 3 autoresearch versions)
    \item Directional rate: 55.23\% (exceeds 51\% profitability threshold)
    \item ML-only reaches $54.9\%$ hit rate; LLM-only comparison uses different metrics (Table~\ref{tab:stockclaw-llm-ml})
    \item VotingClassifier ensemble (XGBoost + LightGBM + ExtraTrees) with 92 features
    \item Full market cycle data (2019--2026): COVID crash, 2021 bull, LUNA/FTX collapse, ETF era
\end{itemize}

\textbf{Critical insight}: LLMs excel at narrative analysis (report quality) but fail at numerical directional prediction. Autoresearch correctly identified this separation: LLM for reports, from-scratch ML for judgment.

\textbf{Caveats.}\quad StockClaw results are from internal evaluation on proprietary data. Hit rate is measured on held-out temporal splits, but we note that a complete assessment would require transaction cost modeling, slippage estimation, and comparison against standard financial baselines (e.g., buy-and-hold). These results demonstrate autoresearch's iterative improvement capability rather than claiming trading profitability.

\section{Case Study 3: Genkidama, BitNet Architecture Optimization}

\subsection{Problem}

Training a BitNet b1.58 language model from scratch requires optimizing a high-dimensional hyperparameter space: architecture (width, depth, FFN ratio, GQA ratio), learning rate schedule (WSD warmup/decay fractions), optimizer settings ($\beta_1$, $\beta_2$, weight decay), and context length.

\subsection{Autoresearch v5: 10-Phase Systematic Search}

Autoresearch decomposed the search into 10 phases, each varying a single hyperparameter group while fixing all others at their current best:

\begin{figure}[h]
\centering
\begin{tikzpicture}
\begin{axis}[
    width=12cm, height=6cm,
    xlabel={Phase},
    ylabel={Validation Loss},
    xtick={1,2,3,4,5,6,7,8,9,10},
    xticklabels={Arch, LR, Method, WSD, $\beta$, FFN, Context, GQA, GradAcc, Final},
    xticklabel style={rotate=30, anchor=east, font=\scriptsize},
    ymin=6.0, ymax=8.2,
    grid=major,
    grid style={dashed, gray!30},
    mark size=2.5pt,
    line width=1pt,
]
\addplot[color=red!70!black, mark=*] coordinates {
    (1, 7.6833) (2, 7.6885) (3, 7.7802) (4, 7.7062)
    (5, 7.7646) (6, 7.9323) (7, 7.1229) (8, 7.9083)
    (9, 7.6552) (10, 6.3990)
};
\end{axis}
\end{tikzpicture}
\caption{Genkidama autoresearch v5: best validation loss per phase. Each value is the best within that phase's sweep, not the cumulative best-so-far across phases. Phase 7 (context length) shows the largest single improvement. Phase 10 validates top-3 configs at 4$\times$ training steps.}
\label{fig:genkidama}
\end{figure}
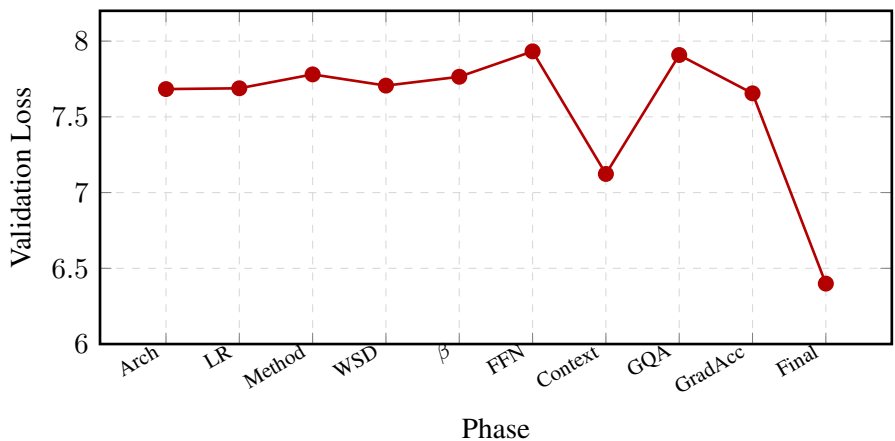

\begin{table}[h]
\centering
\caption{Genkidama autoresearch: phase results and key findings.}
\label{tab:genkidama-phases}
\begin{tabular}{@{}clccp{5cm}@{}}
\toprule
\textbf{Phase} & \textbf{Focus} & \textbf{Configs} & \textbf{Best val\_loss} & \textbf{Key Finding} \\
\midrule
1 & Architecture & 8 & 7.6833 & Wider $>$ Deeper (1024d$\times$12L) \\
2 & Learning rate & 8 & 7.6885 & LR $5\times10^{-4}$--$10^{-3}$ optimal \\
3 & Training method & 6 & 7.7802 & Two-stage WD (matches BitNet 2B4T~\cite{bitnet2b4t}) \\
4 & WSD schedule & 6 & 7.7062 & Warmup 1\%, Decay 15\% \\
5 & $\beta_2$, weight decay & 6 & 7.7646 & $\beta_2{=}0.98 > 0.95$ \\
6 & FFN multiplier & 5 & 7.9323 & 2.67$\times$ FFN optimal \\
7 & Context length & 4 & \textbf{7.1229} & \textbf{Largest effect}: 512$\to$2048 \\
8 & GQA ratio & 5 & 7.9083 & GQA 4:1 competitive \\
9 & Grad accumulation & 3 & 7.6552 & No accumulation best \\
10 & Final validation & 3 & \textbf{6.3990} & 2000-step confirmation \\
\bottomrule
\end{tabular}
\end{table}

\subsection{Results}

\begin{itemize}
    \item 54 configurations across 10 phases, completed in 2.3 hours
    \item Best config: $d{=}1024$, $L{=}12$, $h{=}16$, kv$=4$, ctx$=2048$ (362.9M params)
    \item Validation loss: $7.6833 \rightarrow 6.3990$ ($-16.7\%$) through systematic search
    \item Context length alone accounts for $-7.3\%$ of improvement, the single largest lever
\end{itemize}

\subsection{Scaling to 618M for Full Training}

The autoresearch-optimized config was scaled to 618M parameters (d=1536, L=17, h=16, kv=4, intermediate=4096, ctx=1024) with LLaMA-compatible architecture for bitnet.cpp GGUF export. Pretraining on 3.8B trilingual tokens (EN 38\%, KO 28\%, JA 33\%) has been completed\footnote{As of March 26, 2026: pretraining completed at 480{,}000 steps, best val\_loss 2.3762 (at step 455{,}500), throughput $\sim$6{,}550 tok/s on a single NVIDIA RTX PRO 6000. Successfully exported to HuggingFace format. ORPO (Odds Ratio Preference Optimization~\cite{hong2024orpo}) alignment training is currently in progress using 19{,}566 preference pairs. ORPO unifies SFT and preference optimization in a single loss function, enabling direct conversational alignment from a pretrained base model. Both the pretraining corpus and ORPO preference pairs include synthetic data generated using Claude (Anthropic) as a teacher model. This follows the same teacher distillation paradigm used by Microsoft's Phi series (GPT-4) and Stanford Alpaca (ChatGPT). The long-term goal is for MAGNET's own models to progressively replace the teacher (see Section~\ref{sec:discussion-accessibility}).}.

\begin{table}[h]
\centering
\caption{Genkidama 618M training specifications.}
\label{tab:genkidama-specs}
\begin{tabular}{@{}ll@{}}
\toprule
\textbf{Parameter} & \textbf{Value} \\
\midrule
Architecture & BitNet b1.58 LLaMA (SwiGLU, RMSNorm, GQA, RoPE) \\
Parameters & 618.3M (tied embeddings) \\
Vocabulary & 128{,}256 (Llama 3 tokenizer) \\
Training corpus & 3.8B tokens (trilingual) \\
Token:param ratio & 6.1:1 (data-efficient per Spectra~\cite{kaushal2024}) \\
LR schedule & WSD (Warmup-Stable-Decay) \\
Quantized size & $\sim$122 MB (1.58 bits/weight) \\
Target runtime & bitnet.cpp (CPU-native C++ inference) \\
\bottomrule
\end{tabular}
\end{table}

\section{Cross-Case Analysis: Autoresearch Patterns}

Across three domains, autoresearch exhibits consistent patterns:

\begin{table}[h]
\centering
\caption{Cross-domain autoresearch summary.}
\label{tab:cross-domain}
\begin{tabular}{@{}lcccc@{}}
\toprule
\textbf{Domain} & \textbf{Versions} & \textbf{Configs} & \textbf{Improvement} & \textbf{Duration} \\
\midrule
Zevor (video safety) & 9 & $\sim$5{,}000 & $+5.6\%$ bal\_acc & 2 days \\
StockClaw (crypto) & 3 & $\sim$100 & $+13.9$ pp hit rate & 3 days \\
Genkidama (BitNet LM) & 10 phases & 54 & $-16.7\%$ val\_loss & 2.3 hours \\
\bottomrule
\end{tabular}
\end{table}

\textbf{Consistent findings across domains:}

\begin{enumerate}
    \item \textbf{Error-driven search appears more sample-efficient than uniform exploration in our experiments}: Each version's innovations are targeted at specific failure modes, not uniformly sampled.
    \item \textbf{Hard ceilings were observed and detected in our case studies}: Zevor v8$\to$v9 confirmed 0.9851 as the feature-set ceiling for that dataset. Autoresearch correctly terminated.
    \item \textbf{Architectural pivots are critical}: Zevor's transformer$\to$tree pivot and StockClaw's LLM$\to$ML pivot each provided the largest single improvements. Autoresearch identifies when the current approach has fundamentally wrong assumptions.
    \item \textbf{Ensemble diversity beats single-model optimization}: Zevor (XGB+ET), StockClaw (XGB+LGB+ET), and Genkidama (multi-config validation) all benefit from combining diverse models.
    \item \textbf{Data scale is a separate axis from model optimization}: StockClaw's largest gain came from 14-day$\to$6.5-year data expansion, not model changes. Autoresearch can identify when more data is needed vs. better models.
\end{enumerate}

\section{DiLoCo: Distributed Knowledge Aggregation}
\label{sec:diloco}

The preceding case studies demonstrate that autoresearch can independently produce strong domain specialists on individual nodes. DiLoCo addresses the next challenge: how to combine these independently trained specialists into a collectively stronger model.

\subsection{Protocol}

Following Douillard et al.~\cite{douillard2024}, each node $i$ trains locally for $H$ inner steps using AdamW, then computes pseudo-gradients:
\begin{equation}
    \Delta^{(t)}_i = \theta^{(t-1)} - \theta_i^{(t)}
    \label{eq:diloco-pseudograd}
\end{equation}
where $\theta^{(t-1)}$ is the shared global parameters at the start of the outer step. A coordinator aggregates using an outer optimizer (SGD with Nesterov momentum in the original DiLoCo; we extend this with domain-weighted averaging):
\begin{equation}
    \theta^{(t)} = \theta^{(t-1)} - \eta_{\text{outer}} \sum_{i=1}^{N} w_i \Delta^{(t)}_i
    \label{eq:diloco-aggregate}
\end{equation}
where $w_i$ reflects node $i$'s contribution quality from on-chain evaluation metrics. Note that when $\eta_{\text{outer}} = 1$, $\sum_i w_i = 1$, and $w_i$ reflects local data-size weighting, Eq.~\ref{eq:diloco-aggregate} reduces to a FedAvg-style weighted average update~\cite{mcmahan2017}; the Nesterov momentum and $\eta_{\text{outer}} \neq 1$ are key design choices that significantly improve convergence~\cite{douillard2024}.

\subsection{Adaptations for MAGNET}

\textbf{Heterogeneous specialization}: The designed protocol specifies that nodes intentionally train on different domains, with gradient weights preserving domain expertise.

\textbf{Asynchronous participation}: The protocol allows nodes to submit gradients upon completing $H$ local steps; the coordinator would apply them with staleness weighting.

\textbf{Node tier system}: Nodes are classified into five tiers based on hardware capability, determined via automated hardware profiling and SSD provisioning tests at registration:
\begin{description}
    \item[claude\_session] Autoresearch orchestration + training + storage (3.0$\times$ reward)
    \item[gpu\_compute] GPU training + inference + storage (1.0$\times$)
    \item[storage\_node] SSD storage only, IPFS pin + shard replication (0.3$\times$)
    \item[cpu\_only] CPU inference via bitnet.cpp + storage (0.3$\times$)
    \item[mobile\_light] Relay-assisted mobile/browser inference (0.1$\times$)
\end{description}
Tier multipliers and all reward parameters are defined in a single configuration file (\texttt{config/reward-config.json}), overridable via environment variables, and governance-updateable on-chain via \texttt{Params.tier\_rewards}. Non-SSD nodes are rejected from storage tiers; the provisioning stage includes a 4KB random read/write speed test to filter degraded drives. Each node reports status via periodic heartbeats, enabling decentralized monitoring without centralized log aggregation.

\textbf{BitNet compatibility}: Aggregation operates on float32 latent weights; nodes re-quantize to ternary for inference after each merge.

\subsection{Specialist $\to$ Generalist Cycle}

Each DiLoCo cycle follows a concrete state machine:

\begin{enumerate}
    \item \textbf{Leader election}: Deterministic selection via $\text{leader} = \text{sorted}(\text{eligible})[\text{round} \bmod N]$.
    \item \textbf{Round announcement}: Leader broadcasts \texttt{RoundAnnouncement} (round, $H$ steps, deadline, checkpoint hash) to all eligible peers via gRPC. Leader initializes collection state \emph{before} broadcast to prevent fast-peer signal drops.
    \item \textbf{Local training}: Each node (including leader) trains $H$ local steps. Gradient data is uploaded via dual-path (gRPC direct + IPFS pin).
    \item \textbf{Gradient signaling}: Nodes send \texttt{GradientReadySignal} with hash, IPFS CID, and local loss. Leader self-injects via direct path (bypassing peerBook).
    \item \textbf{Collection}: Leader counts collected gradients. \texttt{needed} = actual accepted peers (not total eligible; unreachable peers excluded). Updated post-broadcast via \texttt{updateRoundNeeded()}.
    \item \textbf{Merge}: When $\text{collected} \ge \text{needed}$, coordinator triggers \texttt{performMerge()} $\to$ Python subprocess executes Nesterov momentum update $\to$ merged checkpoint promoted to \texttt{latest.pt}.
    \item \textbf{Next round}: Coordinator auto-advances; nodes receive new announcement on the merged base.
\end{enumerate}

\section{Robust Decentralized Storage}
\label{sec:storage}

Checkpoint shards and gradient tensors require zero-loss distributed storage across unreliable commodity nodes. MAGNET implements a storage layer combining Reed-Solomon erasure coding, background replication, and health monitoring.

\subsection{Reed-Solomon Erasure Coding}

We use systematic Reed-Solomon coding over GF($2^8$) with primitive polynomial $x^8 + x^4 + x^3 + x^2 + 1$ (identical to RAID-6 and Filecoin). Data is split into $K$ data shards and $M$ parity shards via a Vandermonde encoding matrix:
\begin{equation}
    \mathbf{P}_m[j] = (m+1)^j \quad \text{in GF}(2^8), \quad 0 \le j < K, \; 0 \le m < M
\end{equation}
Any $K$ of $(K+M)$ shards reconstruct the original data via Gauss-Jordan matrix inversion over GF($2^8$). Default configuration: $K{=}4$, $M{=}2$, tolerating 2 simultaneous node failures. Verified across all $\binom{6}{2}{=}15$ two-shard-loss combinations.

\subsection{Storage Tier \& SSD Requirement}

All storage-participating tiers (claude\_session, gpu\_compute, storage\_node, cpu\_only) require SSD. A provisioning stage runs 4KB random read/write tests (1000 blocks) to verify ${\ge}100$~MB/s read, ${\ge}50$~MB/s write. Nodes failing this test are demoted to mobile\_light (inference-only). This filters USB drives and degraded SSDs that would degrade shard replication latency.

\subsection{Background Replication \& Health Monitor}

A daemon runs every 30 seconds, performing: (1)~shard holder liveness checks (stale after 120s), (2)~priority repair queue (critical $\to$ high $\to$ normal), (3)~geo-diverse replication targeting 3--5 replicas per shard. Rate-limited to avoid network saturation. Combined with IPFS pinning for content-addressable persistence, this is designed to approach HDFS-level redundancy on commodity hardware.

\section{P2P Security}
\label{sec:p2p-security}

Decentralized operation requires authentication without a central authority. MAGNET implements a dual-mode authentication system with TLS transport encryption.

\subsection{Node Identity Binding}
Each node derives its identity deterministically from its wallet keypair:
\begin{equation}
    \texttt{nodeId} = \texttt{"node-"} \| \text{SHA-256}(\text{pubkey}_{\text{compressed}})_{[:16\text{hex}]}
\end{equation}
This binds the P2P identity to the on-chain wallet address. Forging a nodeId for a given wallet is computationally infeasible without the corresponding private key (secp256k1). Collision probability becomes non-negligible only around birthday-bound scale ($\approx 2^{32}$ identifiers for the 64-bit truncation), well above practical network sizes.

\subsection{Dual-Mode Authentication}

Control-plane RPCs (peer discovery, DiLoCo round announcements, gradient signals, leader proposals) require signed requests:
\begin{itemize}
    \item \textbf{Wallet mode} (production): secp256k1 compact signature with recovery byte over $\text{SHA-256}(\texttt{nodeId}:\texttt{timestamp})$. Verifiers recover the public key from the signature and enforce nodeId binding.
    \item \textbf{Cluster HMAC mode} (private deployment): HMAC-SHA256 using a shared cluster secret, with timing-safe comparison.
\end{itemize}
Both modes enforce a 5-minute timestamp freshness window. Data-plane RPCs (gradient push/fetch, shard replication) authenticate via gRPC metadata headers carrying the same signature scheme.

\subsection{TLS with Trust-On-First-Use (TOFU)}

All gRPC transport is encrypted via TLS. Nodes auto-generate self-signed certificates; on first contact, a peer's certificate is fetched via raw TLS handshake and stored locally. Subsequent connections verify against the stored certificate. Operators can supply CA-signed certificates via \texttt{MAGNET\_P2P\_CERT/KEY/CA} for production PKI deployments.

\section{On-Chain Incentive Loop}
\label{sec:onchain}

A decentralized training network requires nodes to voluntarily contribute compute, storage, and research effort over extended periods. Without verifiable, tamper-proof reward attribution, rational nodes have no incentive to participate---and the network cannot sustain itself. On-chain incentive mechanisms solve this by making contribution measurement transparent and reward distribution auditable: every node can independently verify that its work was correctly credited. This is not a financial product design but a \emph{systems requirement} for maintaining a self-sustaining distributed compute network.

The mechanisms described in this section are fully implemented as a Cosmos SDK module and validated via unit tests on the HOOTi EVM chain. However, they have not yet been battle-tested on a public mainnet with adversarial participants. Implementation scope includes mining emission, data registration fees, ResearchJob settlement, x402 API revenue distribution, commit-reveal, differentiated slashing, and bootstrap notary consensus. Robustness at scale remains to be demonstrated through mainnet deployment. Accordingly, ``on-chain rewards'' should be understood not as a single token payout but as a collection of revenue loops with distinct state transitions and verification procedures.

\subsection{Contribution Tracking}

Each node's contribution is measured along three axes:

\begin{table}[h]
\centering
\caption{On-chain contribution metrics (implemented, pending mainnet validation).}
\label{tab:contribution}
\begin{tabular}{@{}lll@{}}
\toprule
\textbf{Metric} & \textbf{Measurement} & \textbf{Verification} \\
\midrule
Data quality & Held-out evaluation loss $\Delta$ & Designed for on-chain reproducibility \\
Training compute & Steps $\times$ model size & Checkpoint hash \\
Model quality & Benchmark scores & Evaluator consensus (implemented) \\
\bottomrule
\end{tabular}
\end{table}

\subsection{Reward Distribution}

Base rewards distribute a fixed pool proportionally to measured contribution, scaled by each node's tier multiplier:
\begin{equation}
    R_i = R_{\text{pool}} \times \frac{T_i \, C_i}{\sum_{j=1}^{N} T_j \, C_j}
    \label{eq:reward}
\end{equation}
where $R_{\text{pool}}$ is the total reward pool for the epoch, $T_i$ is the tier multiplier (Section~\ref{sec:diloco}), and $C_i$ is node $i$'s verified contribution score. By construction, $\sum_{i=1}^{N} R_i = R_{\text{pool}}$, guaranteeing zero inflation regardless of node count. In the current implementation, task completions are accumulated during 10-minute epochs; at epoch boundary, the pool is distributed proportionally and recorded both in-memory and on-chain via a dedicated settlement transaction (no per-task reward calculation, preventing double-distribution). Annual emission is capped with a daily on-chain limit to prevent burst exploitation. Fee splits are differentiated by activity type and enforced via on-chain consensus parameters.

\subsection{Commit-Reveal \& Dispute Resolution}

To prevent front-running of evaluation results, contribution proofs follow a two-phase commit-reveal protocol: nodes first submit $H(\text{result} \| \text{nonce})$; after a commit window closes, they reveal the plaintext. Mismatched reveals are slashed.

Dispute resolution uses a \textbf{bootstrap notary panel}: during early network operation, a notary quorum adjudicates challenges. Each notary must post a minimum bond. Differentiated slashing rates are enforced proportional to offense severity.

Data registration follows the same verification philosophy. When a contributor submits training data, the record is not immediately confirmed; it enters a challenge window during which notaries verify data quality and integrity. If no challenges are raised, the record transitions to confirmed status and the registration fee is distributed. Inaccurate or malicious data submissions are subject to differentiated slashing. Thus, data registration revenue is not a simple upload reward but a settlement procedure for verified data contributions.

A burn mechanism creates pressure proportional to network usage. All split ratios are defined in \texttt{Params} and modifiable via on-chain governance proposals.

Design properties: auditable (on-chain records), Sybil-resistant (verified compute), proportional. Tamper-resistance via evaluator consensus is implemented but pending mainnet validation.

\subsection{Incentive Compatibility Analysis}

The mechanism is designed so that rational nodes maximize payoff through honest participation. We formalize this as a cost-benefit inequality analogous to Bitcoin's security model~\cite{nakamoto2008}, where the system is secure when the cost of attack exceeds the expected gain.

\textbf{Definition.} For a simplified stationary analysis, assume a constant expected per-round reward $R_i$ (Eq.~\ref{eq:reward}). Let $B_i$ denote node $i$'s posted bond, $s \in [0.1, 1.0]$ the applicable slashing rate, and $G$ the one-time gain from a successful attack (e.g., claiming reward for fabricated compute). The node is \emph{incentive-compatible} when:

\begin{equation}
    \underbrace{s \cdot B_i + \sum_{t=1}^{\infty} \gamma^t R_i}_{\text{cost of attack (bond loss + forfeited future rewards)}} > \underbrace{G}_{\text{one-time attack gain}}
    \label{eq:ic}
\end{equation}

where $\gamma \in (0,1)$ is a discount factor representing the node's time preference for future rewards.

\textbf{Analysis.} The left side has two components: (1)~the immediate bond loss $s \cdot B_i$, and (2)~the discounted stream of future rewards $\gamma R_i / (1 - \gamma)$ that the node forfeits upon slashing and exclusion (note the sum starts at $t{=}1$, yielding $\gamma R_i/(1-\gamma)$, not $R_i/(1-\gamma)$). As the network matures and $R_i$ grows, the opportunity cost of cheating increases monotonically, strengthening the inequality without requiring bond increases. For fraudulent approval ($s = 1.0$), the attacker loses the entire bond immediately; for lesser offenses ($s = 0.1$ to $0.5$), the future reward forfeiture term dominates for any node with a sufficiently long planning horizon.

\textbf{Sybil resistance.} Creating $k$ identities requires posting $k$ bonds. Under the assumptions of identical tiers, additive contribution scores, a fixed reward pool (Eq.~\ref{eq:reward}), and no minimum-payout cutoffs, splitting into $k$ identities yields $\sum_{j=1}^{k} R_j = R_i$ (the same total reward) while requiring $k \times B$ in bond capital. Identity splitting thus does not increase total reward while increasing capital cost, making Sybil attacks economically unattractive.

\textbf{Limitations.} This analysis assumes: (i)~rational, profit-maximizing nodes (does not model altruistic or adversarial-at-any-cost actors), (ii)~the slashing mechanism correctly identifies misbehavior (dependent on notary panel integrity), and (iii)~bond values are set high enough that $s \cdot B_i$ is non-trivial relative to $G$. Formal verification of these assumptions under adversarial conditions remains future work pending mainnet deployment.

\section{Comparison with Existing Systems}

\begin{table}[h]
\centering
\caption{Feature comparison with existing decentralized AI systems. We compare along MAGNET's four pillars; we acknowledge this selection favors our system and note that competitors may excel on dimensions not shown (e.g., model scale, active participants, production maturity).}
\label{tab:comparison}
\small
\begin{tabular}{@{}lccccp{3.2cm}@{}}
\toprule
\textbf{System} & \textbf{Auto.} & \textbf{Tern.} & \textbf{Merge} & \textbf{Chain} & \textbf{Notes} \\
\midrule
Bittensor & \texttimes & \texttimes & \texttimes & \checkmark & Inference market \\
Gensyn & \texttimes & \texttimes & NoLoCo & In dev. & Compute protocol \\
Prime Intellect & \texttimes & \texttimes & \checkmark & \texttimes & OpenDiLoCo, 10B \\
Nous (DeMo) & \texttimes & \texttimes & \checkmark & \texttimes & Dist. optimizer \\
Together AI & \texttimes & \texttimes & \texttimes & \texttimes & Centralized GPU \\
\textbf{MAGNET} & \checkmark & \checkmark & \checkmark$^\dagger$ & \checkmark$^\ddagger$ & This work \\
\bottomrule
\multicolumn{6}{@{}p{\textwidth}@{}}{\scriptsize $^\dagger$ DiLoCo coordinator, Nesterov outer optimizer, and gradient compression implemented; initial multi-node test completed (drift measured); large-scale validation pending.} \\
\multicolumn{6}{@{}p{\textwidth}@{}}{\scriptsize $^\ddagger$ Implemented and unit-tested on HOOTi EVM chain; not yet battle-tested on public mainnet.}
\end{tabular}
\normalsize
\end{table}

\section{Discussion}

\subsection{From Automated Training to Automated Research}
\label{sec:discussion-accessibility}

TensorFlow and PyTorch democratized model \emph{training}: anyone can run \texttt{model.fit()}. But the research process (identifying what to train, on what data, with what architecture, evaluated how) remains expert-driven. As Karpathy demonstrated with his \texttt{autoresearch} project~\cite{karpathy2026}, the research loop itself can be automated on a single machine. MAGNET extends this by distributing the autonomous research loop across a decentralized network.

A node operator specifies a domain; the system autonomously generates data, trains models, evaluates results, pivots strategies on failure, and produces an optimized expert model. The operator does not need ML expertise or access to proprietary datasets. Training tasks are dispatched to GPU-capable nodes in the network, while CPU-only nodes participate in inference, evaluation, and serving. The research \emph{orchestration} runs on any hardware, but compute-intensive training still requires GPU nodes. MAGNET's hardware accessibility claim applies primarily to the inference and serving layer, not training itself.

This represents an extension along the automation axis: TensorFlow simplified \texttt{model.fit()}; Karpathy's autoresearch~\cite{karpathy2026} automated the single-machine research loop; MAGNET targets the decentralized, multi-node case.

\subsection{Limitations}

The current MAGNET architecture has the following limitations.

\begin{enumerate}
    \item \textbf{Teacher model dependency.} Autoresearch currently relies on Claude for data generation (teacher distillation). As MAGNET's own models improve through DiLoCo merging, they can progressively replace the teacher. The long-term goal is full self-bootstrapping.

    \item \textbf{DiLoCo with heterogeneous data.} Standard DiLoCo assumes similar data distributions across nodes, but MAGNET intentionally trains on different domains per node. The coordinator applies domain-weighted gradient averaging (Section~\ref{sec:diloco}), but full gradient masking for preventing catastrophic forgetting remains future work. Acker et al.~\cite{diloco_drift2025} also found that representation drift during DiLoCo pretraining is irreversible, persisting even after switching to standard DDP training.

    \item \textbf{618M is a pipeline-validation starting point.} A single 618M ternary model cannot match a 70B float16 model in capability. However, MAGNET's goal is not single-model size competition but a \textbf{model network}: large general-purpose models grown via DiLoCo merging, domain specialists (medical, legal, code, etc.), and lightweight decision-only models that differentiate, interconnect, and branch---overcoming individual model size limits at the network level.

    \item \textbf{On-chain module maturity.} The on-chain compute module implements daily emission cap, commit-reveal, differentiated slashing, and bootstrap notary consensus. These are validated via unit tests but not yet battle-tested on public mainnet. As the number of nodes grows, transaction costs and latency may become significant; batch submission and L2 solutions will be necessary at scale.

    \item \textbf{Trust-On-First-Use (TOFU) security model.} P2P TLS uses TOFU for initial peer certificate trust (Section~\ref{sec:p2p-security}). While nodeId is cryptographically bound to wallet keys and all RPCs are signed, the first connection to an unknown peer trusts the presented certificate. For hostile public internet deployments, operators should supply CA-signed certificates. TOFU is appropriate for permissioned/staked networks where economic incentives deter attacks.
\end{enumerate}

\subsection{Broader Implications}

If the architecture validates at scale, MAGNET's design has several implications for the structure of ML research:

\begin{itemize}
    \item \textbf{Lower barriers to participation.} CPU-only nodes can contribute to model serving and evaluation, reducing the hardware floor for participation in model development.
    \item \textbf{Long-tail domain coverage.} Domains too small for centralized providers to serve profitably could be addressed by individual contributors with domain expertise.
    \item \textbf{Auditability.} On-chain contribution records provide a verifiable log of who contributed what, under what conditions.
\end{itemize}

\section{Conclusion}

We presented MAGNET, a four-pillar architecture for decentralized autonomous expert model generation. Through three case studies (Zevor ($0.9287 \to 0.9851$, FN$=0$), StockClaw ($41\% \to 54.9\%$), and Genkidama ($-16.7\%$ val\_loss)), we demonstrated that autoresearch consistently discovers domain-specific optimizations that outperform initial manual configurations through iterative error-driven refinement.

The primary empirical contribution is the autoresearch methodology: an error-driven, multi-version autonomous research pipeline validated across three domains. BitNet b1.58 training is validated through Genkidama (618M parameters, pretraining complete, ORPO alignment in progress). The on-chain incentive module is implemented and unit-tested but awaits mainnet deployment. DiLoCo distributed merging remains a design for future work. While alternatives exist for individual components (4-bit quantization for BitNet, FedAvg for DiLoCo, reputation systems for incentives), their specific combination targets the properties required by decentralized AI research: hardware accessibility, communication efficiency, and verifiable incentives.

Our work builds directly on Karpathy's autoresearch concept~\cite{karpathy2026}, extending it from a single-machine script to a decentralized, multi-version pipeline. The case studies suggest that autonomous, error-driven research iteration is a viable methodology across diverse ML domains. Whether the full four-pillar architecture can deliver on its design goals at scale remains an open empirical question that we plan to address through progressive deployment.

\section*{Acknowledgments}

We thank Alex Jones for the open-source \texttt{llmfit} tool~\cite{llmfit2025}, which provides automated hardware profiling and LLM compatibility assessment for MAGNET's node tier classification system.

\paragraph{Use of AI tools.} Claude (Anthropic) was used as a teacher model for generating synthetic training data and ORPO preference pairs for Genkidama, as described in Section~8. Claude Code was used as an assistive tool during manuscript preparation (editing, formatting, and proofreading). The authors take full responsibility for the accuracy and integrity of all content.


\end{document}